\documentclass[a4paper, 10pt, conference]{ieeeconf}      
\usepackage{FG2025}

\FGfinalcopy 

\IEEEoverridecommandlockouts 
\usepackage{graphicx} 
\usepackage{epsfig} 
\usepackage{amsmath} 
\usepackage{amssymb}  

\usepackage{booktabs}
\usepackage{diagbox}
\usepackage{multirow}
\usepackage{color}
\usepackage{subcaption}
\usepackage{array}
\usepackage{makecell}
\usepackage{float}

\makeatletter  
\let\NAT@parse\undefined 
\makeatother
\usepackage[pagebackref,breaklinks,colorlinks,citecolor=teal]{hyperref} 

\def\FGPaperID{152} 

\title{\LARGE \bf
Trajectory-guided Motion Perception for Facial Expression Quality Assessment in Neurological Disorders}

\author{ %
\parbox{16cm}{\centering
   {\large Shuchao Duan\textsuperscript{1}}, %
   {\large Amirhossein Dadashzadeh\textsuperscript{1}}, %
   {\large Alan Whone\textsuperscript{2}}, %
    {\large Majid Mirmehdi\textsuperscript{1}}\\
    \textsuperscript{1}School of Computer Science \hspace{0.2cm} \textsuperscript{2}Translational Health Sciences \\
    University of Bristol, UK\\
    {\tt\small \{shuchao.duan, a.dadashzadeh,  alan.whone, m.mirmehdi\}@bristol.ac.uk}
}}

\begin{document}

\ifFGfinal
\thispagestyle{plain}
\pagestyle{plain} 
\else
\author{Anonymous FG2025 submission\\ Paper ID \FGPaperID \\}
\pagestyle{plain}
\fi

\maketitle
\thispagestyle{plain}

\begin{abstract}
Automated facial expression quality assessment (FEQA) in neurological disorders is critical for enhancing diagnostic accuracy and improving patient care, yet effectively capturing the subtle motions and nuances of facial muscle movements remains a challenge. We propose to analyse facial landmark trajectories, a compact yet informative representation, that encodes these subtle motions from a high-level structural perspective. Hence, we introduce Trajectory-guided Motion Perception Transformer (TraMP-Former), a novel FEQA framework
that fuses landmark trajectory features for fine-grained motion capture with visual semantic cues from RGB frames, ultimately regressing the combined features into a quality score. Extensive experiments demonstrate that TraMP-Former achieves new state-of-the-art performance on benchmark datasets with neurological disorders, including PFED5 ($\uparrow6.51\%$) and an augmented Toronto NeuroFace ($\uparrow7.62\%$). Our ablation studies further validate the efficiency  and effectiveness of landmark trajectories in FEQA. Our code is available at \url{https://github.com/shuchaoduan/TraMP-Former}.


\end{abstract}

\section{INTRODUCTION}

{Facial expression quality assessment (FEQA) in videos 
enables objective symptom evaluation and continuous patient monitoring, offering transformative potential for assessing the severity of neurological disorders, such as Parkinson’s disease (PD) \cite{duan2023qafe}, amyotrophic lateral sclerosis (ALS) \cite{torontoNeuroface}, and other pain-related diseases \cite{pain_dataset_2011}. By providing non-invasive, real-time insights into patient conditions, FEQA has the potential to greatly improve clinical decision-making and alleviate healthcare workloads.}

{Facial expressions include typical gestures that can convey emotions, such as happiness, sadness, anger, surprise, disgust, and fear, while more particular facial actions or expressions, such as `clenching teeth', `raising eyebrows', `blowing a kiss', amongst others, are often used in the clinic as valuable indicators for assessing the severity of certain neurological conditions.}
Current methods for FEQA in healthcare primarily rely on RGB video input, and analyse its spatio-temporal features  \cite{distanceordering2021,rajasekhar2021deep,de2021mdn,huang2023auto,de2024facial,liao2024sequence, liu2024hierarchical,wu2024global}. 
To increase accuracy, 
 while reducing computational overheads, 
others have explored {additional} modalities {to either complement \cite{kachele2017adaptive,thiam2020two,semwal2021computer,Hou2021AM2,dynamicfeaturesPD,moshkova2022assessment,gomez2023exploring,duan2023qafe,razzouki2024early} or replace \cite{xu2020AU,moshkova2020facial,Szczapa_2022_trajec,ipapo2023clinical,oliveira2024video}} RGB features with more focused features, such as optical flow \cite{thiam2020two,razzouki2024early}, action units (AUs) \cite{xu2020AU,moshkova2020facial,moshkova2022assessment,gomez2023exploring,oliveira2024video}, and facial landmarks \cite{kachele2017adaptive,Hou2021AM2,semwal2021computer,dynamicfeaturesPD,Szczapa_2022_trajec,ipapo2023clinical,duan2023qafe}. 

Among these studies, the majority have targeted specific expressions, such as  depression or (shoulder) pain, to evaluate health states, and some others \cite{moshkova2020facial,Hou2021AM2,dynamicfeaturesPD,gomez2023exploring,huang2023auto,oliveira2024video,razzouki2024early} have simply attempted early detection of neurological disorders. However, our work aims to assess symptom severity and progression in neurological conditions through fine-grained analysis of subtle facial muscle movements within expressions. This task is particularly challenging, as it requires measuring the severity and subtlety of a range of irregular expressions. Such analysis involves extracting features like movement amplitude, facial muscle symmetry, and dynamic progression, all of which reflect the disease’s impact on facial function.  Only a few, such as \cite{moshkova2022assessment,ipapo2023clinical,duan2023qafe},
have examined facial expressions in neurological disorders to this extent, {and further investigation for more accurate measurement is still necessary.} 

\begin{figure}[t]
\captionsetup[subfigure]{justification=centering}
    \centering
    \begin{subfigure}{0.48\linewidth}
        \includegraphics[width=\linewidth]{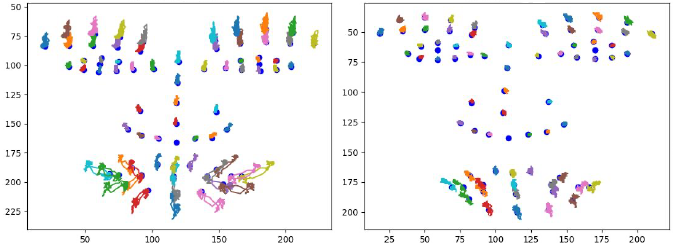}
        \caption{\small{Trajectories projected in 2D}}
        \label{fig:short-a}
    \end{subfigure}
    \begin{subfigure}{0.5\linewidth}
        \includegraphics[width=\linewidth]{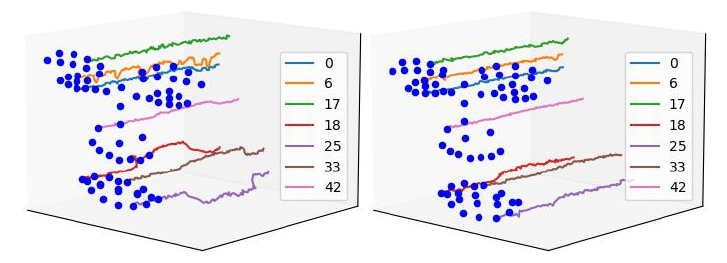}
        \caption{\small{Example trajectories in 3D}}
        \label{fig:short-b}
    \end{subfigure}
    \hfill
    \begin{subfigure}{\linewidth}
        \includegraphics[width=\linewidth]{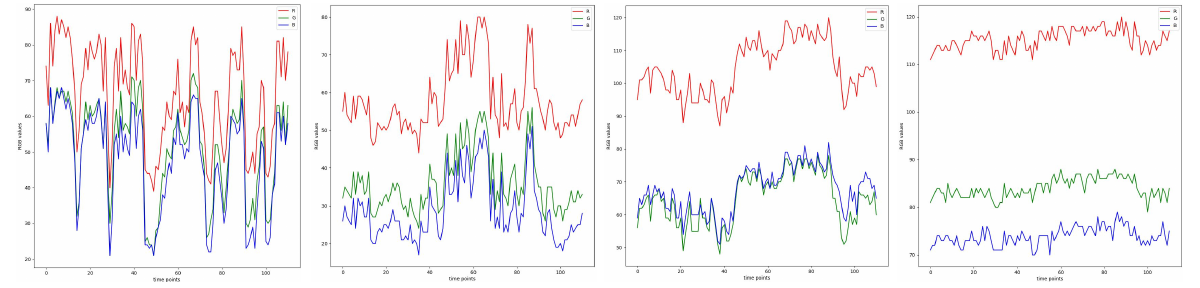}
        \caption{\small{Variation of RGB values over time on example landmarks of the healthy subject (Landmarks 0, 17, 25, and 42)}.}
        \label{fig:short-c}
    \end{subfigure}
    
\caption{{{\bf Landmark trajectories during a clench-teeth action -} In (a) and (b), the left panel is for a healthy subject and the right for a PD patient. The healthy case presents apparent variations in the mouth region with subtle motions in the eyebrows and nose areas, whereas the severe PD case exhibits significantly diminished movements in the mouth and little to no motion in the nose or eyebrows.
In (c), RGB value variations are shown for 4 example landmarks across time for the healthy subject. Overall in this work, a trajectory comprises position and pixel value for each point.}}


\label{fig:modality_example}
\vspace*{-3mm}
\end{figure}

{To efficiently capture the nuances of neurological facial expressions for FEQA, we 
propose using trajectories 
of significant points on the face, {where a trajectory is defined as} the motion path of landmarks 
and their associated RGB pixel values}. We hypothesise that such characterisation has the ability to encode subtle facial muscle movements more effectively, offering valuable insights into keypoint dynamics that could enhance the spatio-temporal analysis of expressions.  
{Figures \ref{fig:short-a} and \ref{fig:short-b} visualise {the spatial position of} landmark trajectories of two subjects (healthy on the left panel and PD patient on the right panel  in each case) during the clinical assessment test of `clenching teeth' in 2D and 3D spaces. The 2D projection highlights that these keypoint movements occur within confined regions whereas the 3D depicts the landmark trajectories over time. 
{For the healthy subject, we observe pronounced activity in the mouth region, along with subtle movements in adjacent areas like the eyebrows and nose. In contrast, the PD subject exhibits severe symptoms, with significantly reduced movement in the mouth region and a near absence of motion around the eyebrows and nose.} 
{Fig. \ref{fig:short-c} shows the variation of the RGB values associated with the respective trajectory for several example landmarks.} 
}


{By their nature, landmarks represent key positions on the face for efficient feature representation, but we must not neglect other `landmark-less' facial regions that may still be pertinent to our task, such as the cheeks and forehead. Hence we incorporate RGB data from within the entire facial region 
to provide visual context and enable a more comprehensive semantic understanding of facial expressions.} 


In this paper, we propose a {Trajectory-guided Motion Perception Transformer (TraMP-Former)},  a {novel FEQA}
framework that combines {semantically rich high level keypoint coordinates from 2D facial landmark trajectories with raw low-level visual pixels} from RGB frames. Specifically, we employ SkateFormer \cite{do2024skateformer} to encode the spatio-temporal motion patterns of the facial landmark trajectories, while Former-DFER \cite{formerDFER} 
is utilised as the RGB encoder to capture visual semantic features. These two features are then fused within several Trajectory-guided Motion Perception (TraMP) Blocks. Here, RGB features, rich in visual details (e.g., color, texture), are used as the query in TraMP Blocks to enhance structure and motion information from the trajectory features relevant to the visual context. Within these TraMP Blocks, only the RGB stream features are updated, while the trajectory stream features remain unchanged. 

Our key contributions are: 
(i) We advocate facial landmark trajectories as a compact, yet informative, representation for capturing subtle facial motions in neurological disorders; 
(ii) We propose TraMP-Former as a model that integrates landmark trajectories and RGB frames to encode both motions and visual semantics; (iii) We present extensive experiments and ablation studies, demonstrating state-of-the-art (SOTA) performance on PFED5 and {an augmented} Toronto NeuroFace dataset, with an average improvement {in Spearman's Rank Correlation} of 
6.51\% and 7.62\%, respectively.


\section{RELATED WORKS}
\label{sec:literature}
We briefly review recent literature relevant to our work, focusing on FEQA in healthcare and modelling spatio-temporal features from trajectories. 

\noindent {\bf Facial Expression Quality Assessment in Healthcare --}
Numerous studies have explored assessing patients' conditions using specific or limited expressions, such as those associated with depression and pain \cite{kachele2017adaptive,thiam2020two,xu2020AU,distanceordering2021,rajasekhar2021deep,de2021mdn,semwal2021computer,Szczapa_2022_trajec,de2024facial,liao2024sequence, liu2024hierarchical,wu2024global}.
For example, Thiam et al. \cite{thiam2020two} proposed a two-stream hybrid CNN-BiLSTM, for analysing pain expressions. Their method combined optical flow and motion history  to capture spatiotemporal information fusing them at the decision level.
Xu et al. \cite{xu2020AU} estimated clip-level shoulder pain based on clip-level pain annotations and  frame-level labeled AUs as complementary features.
{To the best of our knowledge, only \cite{Szczapa_2022_trajec} has explored landmark trajectories for FEQA (for pain estimation). They proposed representing trajectory patterns on the Riemann manifold using the Gram matrix, computed from landmark coordinates and velocities. Support vector regression was then applied to predict pain levels from the resulting similarity matrix, with the final score determined through a late fusion strategy.}

{There are very few key works on fine-grained analysis of facial expressions to assess disease severity  in neurological disorders \cite{moshkova2022assessment,ipapo2023clinical,duan2023qafe}.}
{For example, Ipapo et al. \cite{ipapo2023clinical} used facial landmarks to extract statistical features for analysing expressions in ALS and stroke patients. Different features were tailored to specific quantifiable categories, such as range of motion (ROM), speed, and symmetry, which were used in a Random Forest Classifier to predict severity scores.
Duan et al. \cite{duan2023qafe} proposed QAFE-Net for PD expression analysis by combining facial landmark heatmaps with RGB data. This captured spatial relationships between neighbouring points to enable joint consideration of local and global facial regions.} 
{These methods predominantly explored the spatial relationships of facial muscle movements rather than their temporal evolution. Our aim here is to model fine-grained motions from spatial and temporal perspectives simultaneously.}




\begin{figure*}[t]
    \centering
\includegraphics[width=\linewidth]{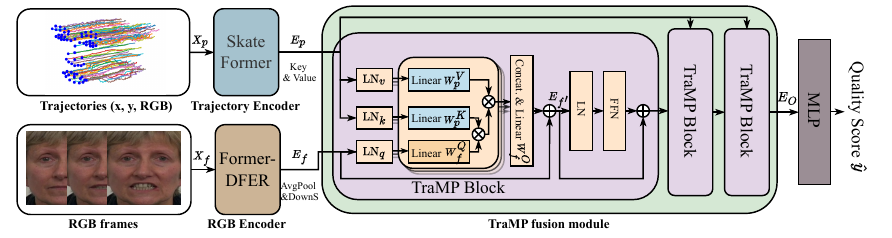}
    \caption{{{\bf Proposed TraMP-Former pipeline}. {Input trajectories, comprising 2D landmark positions with associated RGB values over time and RGB frames are encoded via the Trajectory Encoder and RGB Encoder to produce trajectory features ($E_{p}$) and RGB features ($E_{f}$), respectively.  $E_{f}$ are average pooled and downsampled to align with $E_{p}$. These are fused in the TraMP fusion module, where $E_{p}$ serves as key and value, and $E_{f}$ as the query, to generate the final 1D representation $E_{O}$, which is passed through an MLP to predict the clinical score $\hat{y}$.}} 
    }
    \label{fig:main}
    \vspace*{-3mm}
\end{figure*}

\noindent {\bf Spatio-Temporal Modelling with Trajectories --}
{Psychophysical} studies have demonstrated that dynamic information encoded within trajectories is sufficient for accurate recognition of actions \cite{johansson1973visual} and facial emotions \cite{bidet2020recognition,roberti2024single}. 
In early works \cite{Wang2011ActionRB}\cite{wang2013action}, dense points were sampled from each frame and tracked using optical flow as dense trajectories for action recognition.  Afshar et al. \cite{afshar2016facial} extended this to dynamic facial expression recognition (DFER).
Trajectories formed by skeleton joints have also emerged as a popular modality, offering advantages like insensitivity to background and viewpoint variations for {human action analysis, e.g., \cite{wang2016action,liu2020trajectorycnn,liu2021motion,zhong2022spatio,usmani2023skeleton}.} 
As Transformers excel at capturing contextual relationships through their attention mechanism, they have been explored for analysing trajectories in action analysis \cite{rao2023transg,zheng2024spatio,rao2025motif}. 
For example, Zheng et al. \cite{zheng2024spatio} proposed the Joint Trajectory GraphFormer to learn the spatio-temporal relationships between joint trajectories within a unified Joint Trajectory Graph for action recognition. 
Their model computed scores that quantified the interactions between joint trajectories, enabling it to distinguish similar actions, recognise complex movements, and also capture temporal dependencies. {This attention-based analysis of interactions between trajectories forms the basis of our proposed approach to 
facial expression quality assessment.
}

{We note that Do et al. \cite{do2024skateformer} developed SkateFormer without predefining the topology of skeleton joints, and achieved SOTA performance in modeling skeletal-temporal relations for action recognition while offering a more flexible framework for directly processing input data 
as key points tracked over time. Given the structural and temporal similarities between facial landmarks and skeleton joints - both capturing high-level representations of structure and motion - we apply SkateFormer as the backbone for encoding dynamic patterns from landmark trajectories (for details see Section \ref{sec:approach}).}
\section{PROPOSED METHOD}
\label{sec:approach}

{Fig. \ref{fig:main} illustrates our proposed TraMP-Former framework.  Given video clip $X$ of a specific facial expression containing $T$ frames,
and its associated quality score $y$, the input to TraMP-Former comprises (i) trajectories of $P$ landmarks,  $X_{p} \in \mathbb{R}^{T_{p} \times P \times C_{p}}$, where $T_{p}$ is the length of trajectories and $C_{p}$ is the input channel dimension, and each landmark trajectory is represented as a sequence of 2D position with associated RGB values, denoted as $(x, y, RGB)$ (i.e., $C_p=5$) at each time point,} 
and (ii) a set of RGB frames $X_{f} \in \mathbb{R}^{T_{f} \times H \times W \times 3}$, consisting of $T_{f}$ frames at height  $H$  and width  $W$, sampled from $X$. 

{TraMP-Former has four key components: trajectory encoder $\mathcal{P}$, RGB encoder $\mathcal{R}$,
TraMP fusion module, 
and regression head $\mathrm{MLP}$.} The objective is to learn the parameters $\Theta$ 
so that it can capture informative features for predicting the  quality score $\hat{y}$, i.e.,
\begin{equation}
    \hat{y} = \mathrm{MLP}_{ \Theta}\{\mathcal{P}(X_{p}) \circ \mathcal{R}(X_{f}) \},
\end{equation}
where $\circ$ represents the TraMP fusion processing.

\noindent {\bf Trajectory Stream --}
{To encode the subtle variations of expressions efficiently,
we wish to exploit temporal} facial landmark trajectories, consisting of 2D facial landmarks and associated RGB values over time.

The SkateFormer architecture \cite{do2024skateformer} employs a skeletal joint and frame partitioning strategy to classify skeletal-temporal relationships into local and global motions across sets of neighboring and distant joints. This approach enables efficient self-attention within each partition, effectively capturing both detailed and overarching motion dynamics. Inspired by this design, {we use SkateFormer by replacing its skeleton joint coordinates input with our landmark positions and their associated RGB pixel values per frame to model landmark-temporal relationships $[T_{p}, P]$.} 
%

\begin{figure}[t]
    \centering
    \includegraphics[width=0.88\linewidth]{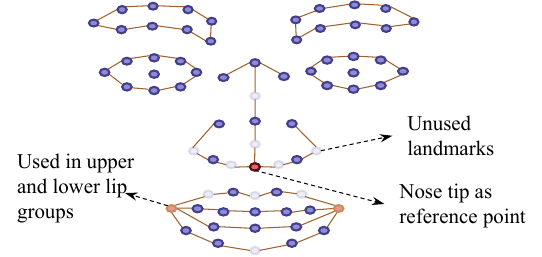}
    \\
    \vspace{0.2cm}
    \includegraphics[width=0.7\linewidth]{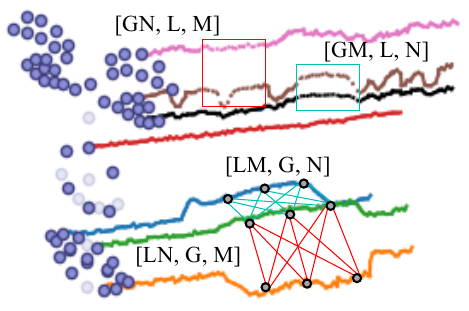}
    \caption{{{\bf Landmarks and their grouping -} (top) The $P$ landmarks are split into $M=7$ groups of $N=9$.
    (bottom) Example landmark-temporal partitioning: {The pink and brown dotted trajectory segments in the red box represent the spatio-temporal relationship between two trajectories from different groups within a local temporal segment, while the  brown  and black dotted trajectory segments in the light green box indicate the spatio-temporal relationship between two trajectories from the same group (i.e., the left eye) within a local temporal segment.}
    The red cross-lines depict the correlation between two landmarks from different groups within global motion, and light green cross-lines represent the correlation between two landmarks from the same group (i.e., upper lip) within global motion.}}
    \label{fig:split_example}
   \vspace*{-3mm}
\end{figure}

{We divide} these landmark-temporal
relationships $[T_{p}, P]$ 
into four partition types, as shown in Fig. \ref{fig:split_example} (bottom). 
To encode motion semantics, we define two temporal axes,  local motion $t_{g}^{local}$ and global motion $t_{l}^{global}$ as   
\begin{equation}
    t_{g}^{local} = [(g-1)L+1,(g-1)L+2,...,gL],
\end{equation}
\begin{equation}
    t_{l}^{global} = [l,l+L,...,l+(G-1)L],
\end{equation}
where $g =1,2,...,G$ with total $G$ number of $ t_{l}^{global}$ elements, i.e., $G$ represents the number of temporal segments, and $l =1,2,...,L$ with total $L$ number of $ t_{g}^{local}$ elements, i.e., $L$ is the number of frames per $ t_{g}^{local}$ segment. Here, $t_{g}^{local}$ is a contiguous segment of time indices for capturing local motion, while $t_{l}^{global}$ is an $L$-strided sparse time axis for capturing global motion. 
Similarly, $N$ landmarks within the same facial region are defined as neighboring landmarks, while those from $M$ different regions are considered distant landmarks. Thus, the total number of frames is $T_{p} = G \times L$, and the total number of landmarks is $P = M \times N$.

The four partition types are:
(i) Neighboring landmarks with local motion $[G \times M, L, N]$, (ii) Distant landmarks with local motion $[G \times N, L, M]$, (iii) Neighboring landmarks with global motion $[L \times M, G, N]$, and (iv) Distant landmarks with global motion $[L \times N, G, M]$. 
We partition facial landmarks  into  $M = 7$  regions,  i.e., left eye, right eye, nose, left brow, right brow, upper lip, and bottom lip, each containing  $N=9$  landmarks, resulting in a total of $P = 63$ landmarks (see Fig. \ref{fig:split_example}).  This partitioning 
ensures consistency in the number of landmarks per region. {The nose and lips regions contain more landmarks compared to other facial areas. 
We retain only representative landmarks in the nose region to reduce redundancy, as facial expressions rarely affect this area and its dense coverage contributes little additional information.}
{For the lips, we focus on landmarks along the inner lip contour, as they more effectively capture subtle changes -- such as the separation between the upper and lower lips (e.g., during smiling) -- compared to those on the outer areas of the lips.}
The temporal sequence $T_{p}$ is divided into  $G$  segments,  each containing  $L=8$  frames. 

Since video clips may vary in length, {we standardise all trajectories to a fixed temporal length $T_{p}$, which approximates the average clip length in the dataset and satisfies SkateFormer's partitioning requirements. Trajectories shorter than  $T_{p}$  are extended by looping the video until the desired length is reached, while longer ones are truncated to retain their latter portion, 
to account for delayed expression onset.}
The trajectories $X_{p}$ are processed by the Trajectory Encoder, engaging the defined landmark-temporal relationships to extract 1D trajectory feature vectors $E_{p}$. 


\noindent {\bf RGB Stream --}
{To capture the appearance details in non-landmark regions,
we also process the full RGB clip using the Former-DFER \cite{formerDFER} backbone.}
This choice allows for direct comparison with \cite{duan2023qafe} (as shown later in Section \ref{sec:sota_compare}), 
while capitalising the model’s ability to enhance performance solely through architecture design, without relying on complex techniques such as loss functions or auxiliary tasks. Each RGB input clip $X_{f}$
is 
divided into  $n$  subclips to meet the Former-DFER input requirement of 16 frames per subclip, ensuring  $T_{f} = n \times 16$. {If the length of original clip is shorter than $T_{f}$, it is looped back to the beginning for padding.}
{After these $n$ subclips are processed by the RGB Encoder to extract RGB features,  average pooling is applied 
followed by downsampling, to obtain the 1D RGB feature vectors $E_{f}$.} These operations align $E_f$ with $E_p$, ensuring that the query and key dimensions later in the TraMP fusion module meet the requirements for the attention process.

\noindent {\bf Trajectory-guided Motion Peception Fusion --}
Effectively using the complementary characteristics of the above two modalities is crucial for capturing subtle movement features efficiently from the spatio-temporal space. 


We fuse spatio-temporal features extracted from the trajectory stream ($E_{p} \in \mathbb{R}^{1 \times D}$) and the RGB stream ($E_{f} \in \mathbb{R}^{1 \times D}$)  in our TraMP fusion module, consisting of three TraMP blocks. 
Only the RGB features in the RGB stream are updated during sequential block processing, while the trajectory features remain unchanged to preserve the motion dynamics captured by the trajectory data. These blocks enhance the perception of subtle motions on top of RGB features by utilising motion details from trajectory data.

Each TraMP Block, shown in Fig. \ref{fig:main}, is based on a transformer structure, incorporating a multi-head cross-attention layer and a feedforward network (FFN) layer. Unlike traditional transformer blocks, the multi-head scaled dot-product attention in TraMP is cross-modal: the RGB stream features $E_{f}$ serve as queries, while trajectory stream features $E_{p}$, containing richer motion information, act as keys and values. We analyse this configuration and compare it with other fusion strategies in Section \ref{sec:ablation}. The cross-attention is 
\begin{equation}
    E_{f'}^{i} = E_{f}^{i} + \mathrm{MultiHead}^{i} (E_{p}, E_{f}^{i}),
\end{equation}
where $i \in \{1,2,3\}$ denotes $i^{th}$ TraMP Block and 
$E_{f}^{1} = E_{f}$. 
Note, before applying cross-attention, features are normalised {to obtain
query, key, and value, and then they are} 
split into  $z$  channels for parallel attention processing as 
\begin{equation}
    [E_{q1}^{i}, E_{q2}^{i}, ..., E_{qz}^{i}] = \mathrm{Split}(\mathrm{LN}_{q}(E_{f}^{i})),
\end{equation}
\begin{equation}
    [E_{k1}^{i}, E_{k2}^{i}, ..., E_{kz}^{i}] = \mathrm{Split}(\mathrm{LN}_{k}(E_{p})),
\end{equation}
\begin{equation}
    [E_{v1}^{i}, E_{v2}^{i}, ..., E_{vz}^{i}] = \mathrm{Split}(\mathrm{LN}_{v}(E_{p})),
\end{equation}
{where $ \mathrm{LN}$  denotes layer normalisation.} For each attention head  $h $, the normalised and split features $E_{qh}^{i}$, $E_{kh}^{i}$, and $E_{vh}^{i}$ are linearly projected into query, key, and value representations
through the equations
$Q_{fh}^{i} = E_{qh}^{i} \cdot W^{Q_{i}}_{f}$,
$K_{ph}^{i} = E_{kh}^{i} \cdot W^{K_{i}}_{p}$,
and $V_{ph}^{i} = E_{vh}^{i} \cdot W^{V_{i}}_{p}$.
The cross-attention process for each head  $h$  is then computed as 
\begin{equation}
    \mathrm{Attn}_{h}^{i} = \mathrm{Softmax} (\frac{Q_{fh}^{i} \cdot K^{i\top}_{ph}}{\sqrt{d_{k}}} ) \cdot   V_{ph}^{i},
\end{equation}
where $h \in \{1,..,z\}$,  $\frac{1}{\sqrt{d_{k}}}$ is used to prevent gradient vanishing during training, {and we employ  $z = 8$  attention heads}.
The outputs of all heads in block $i$ are then concatenated and linearly transformed by $W^{O_{i}}_{f}$.

{Following the attention layer, the FFN refines the fused features to obtain $E_{f}$ as input to the next TraMP block, i.e.,
\begin{equation}
   E_{f} = E_{f'}^{i} + \mathrm{FFN}(\mathrm{LN}(E_{f'}^{i})) ~ \forall i, 
\end{equation}
where for the last TraMP block we rename $E_{f}$  as $E_{O}$.}
The FFN contains two fully connected layers with a GELU activation in between.
Finally, $E_{O} $ is passed through an MLP to predict the facial expression quality score  $\hat{y}$ . 

\noindent {\bf Optimisation --}
To effectively supervise the quality assessment process {and mitigate imbalanced regression in the datasets}, we adopt the Batch-based Monte-Carlo (BMC) loss function \cite{balancedMSE_cvpr},
\begin{equation}
    \mathcal{L}_{\mathrm {BMC}   }   =-log\frac{exp(-\left \|  \hat{ y}- y  \right \|_{2}^{2} /\tau  )}{ {\textstyle \sum_{ y^{'}\in  Y  }^{}} exp(-\left \|  \hat{ y}- y^{'}   \right \|_{2}^{2} /\tau  )} ,
\label{eq:bmc}
\end{equation}
where $\tau=2\sigma^{2}_{noise}=2$ is a temperature coefficient, with $\sigma_{noise} =1$, $Y$ represents the set of quality scores, and the loss encourages the predicted score $\hat{y}$ to closely match the ground truth $y$ while mitigating the impact of imbalanced data distributions.

\section{EXPERIMENTS}  
\label{sec:exp_setting}


\subsection{Implementation Details}
\label{sec:imple_details}

\begin{figure} [t]
    \centering
\includegraphics[width=\linewidth]{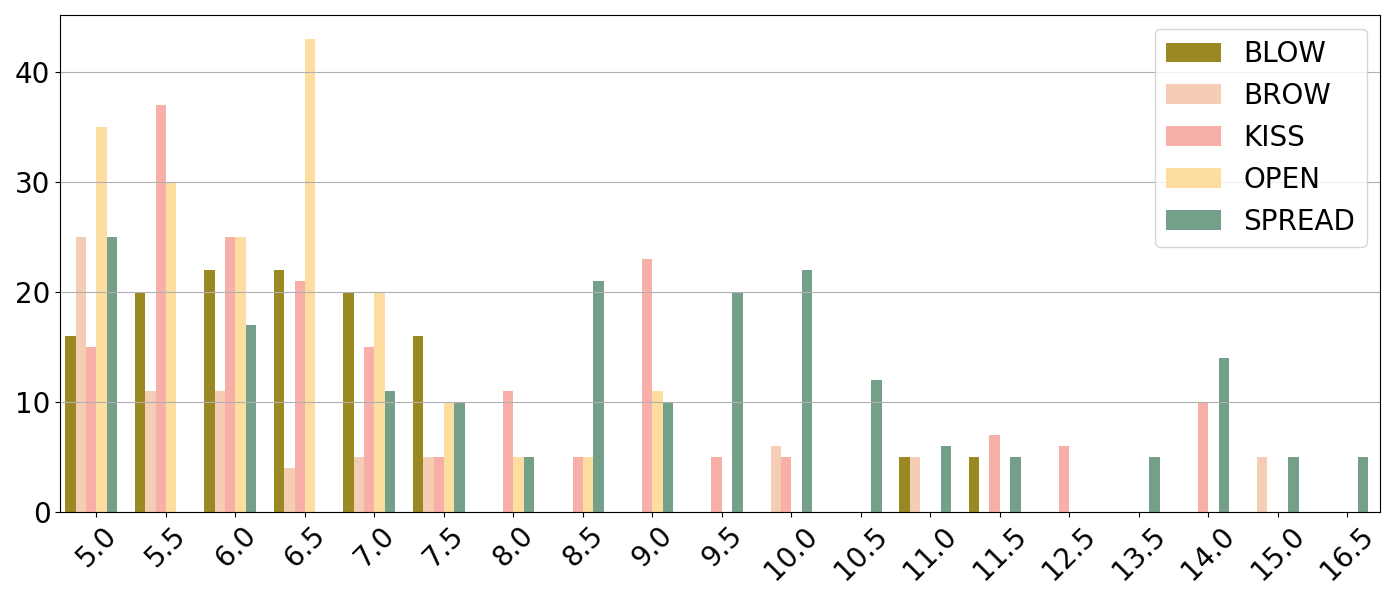}
    \caption{{Score distribution for each action in the augmented Toronto NeuroFace dataset. The x-axis represents the average overall score from two raters and the y-axis indicates the frequency of each score.}}
    \label{fig:neuro_score}
    \vspace*{-3mm}
\end{figure}

 \noindent {\bf Datasets --}
{We evaluate our approach on two publicly available facial expression datasets collected in clinical settings. The \textbf{PFED5} dataset \cite{duan2023qafe} comprises 2,811 videos at 25 fps of 41 subjects with varying PD severity, performing five distinct facial expressions: {`Sit at rest', `Smile', `Frown', `Squeeze eyes', and `Clench teeth'.}
Video lengths range from 7 to 260 frames, and each is annotated with MDS-UPDRS scores \cite{mds-updrs2008}, spanning 0 (normal) to 4 (severe). We use the same training and test split as \cite{duan2023qafe}, i.e., 30 and 11 subjects, respectively.} 
The \textbf{Toronto NeuroFace Dataset} \cite{torontoNeuroface} includes 36 subjects, divided into three groups: 11 with ALS, 14 with stroke, and 11 healthy controls. Each subject performs six distinct non-speech tasks - `BLOW', `KISS', `OPEN', `SPREAD', `BIGSMILE'\footnote{Due to its limited size 
, the BIGSMILE task is excluded from this study.}, and `BROW' - each repeated around five times and recorded at around 50 fps, with two raters providing scores across five aspects: speed, symmetry, ROM, variability, and fatigue. Ratings are based on a 5-point Likert scale, along with an overall score. 
Following \cite{torontoNeuroface}, the averaged overall scores from both raters are used as ground truth. To address the impact of the dataset’s limited size {(see Table \ref{tab:toronto_details}),}
{we manually split each video into subclips,} with each corresponding to a single task repetition under the assumption of consistent symptom severity across repetitions.  Details of our {augmented}
dataset are summarised in Table \ref{tab:toronto_details}. The dataset exhibits an imbalanced score distribution (see Fig. \ref{fig:neuro_score}), similar to that in PFED5 \cite{duan2023qafe}. 
We perform 3-fold cross-validation, {dividing the 36 subjects into 3 folds of 12 each, ensuring that each fold is proportionally representative of the three groups.} 

\begin{table}
\setlength{\tabcolsep}{1.48pt} 
\centering
\scriptsize
\caption{{{\bf Details of our augmented Toronto NeuroFace dataset.} {m/M/A} 
represent the minimum, maximum, and average clip lengths for each expression, respectively. \#Original refers to the number of original samples, each of which may contain multiple repetitions of the expression. \#Current represents the number of samples after splitting, where each sample contains only a single repetition of the expression. 
}}
\label{tab:toronto_details}
\begin{tabular}{l|cccccc} 
\hline
 & \textbf{BLOW} & \textbf{KISS} & \textbf{BROW} & \textbf{OPEN} & \textbf{SPREAD} & \textbf{All} \\ 
\hline
\textbf{\#Original} & 24 & 36 & 15 & 36 & 36 & 147 \\
\textbf{\#Current} & 126 & 190 & 77 & 184 & 193 & 770 \\
\textbf{m/M/A} & 38/245/103 & 34/314/94 & 35/263/87 & 42/306/108 & 33/490/116 & 33/490/104 \\
\hline
\end{tabular}
\end{table}

 \noindent \textbf{Evaluation Metric --} 
 Similarly to the SOTA action quality assessment (AQA) works, such as \cite{2020USDL, CoRe,aqa_tpt}, we evaluate our model performance using Spearman's rank correlation $\rho$ 
 for each expression, i.e. 
\begin{equation}
    \rho = \frac{ {\textstyle \sum_{i}^{}}(\mu_{i}-\bar{\mu} )(\nu_{i}-\bar{\nu} ) }{\sqrt{{\textstyle \sum_{i}^{}}(\mu_{i}-\bar{\mu} )^{2}{\textstyle \sum_{i}^{}} (\nu_{i}-\bar{\nu} )^{2} } } ,
\end{equation}
where $\mu$ and $\nu$ indicate the scores ranking for each sample of two series, respectively.

 \noindent {\bf Data Preparation --} 
Whilst various {face and landmark detection} methods, such as \cite{retinaface2020, guo2021SCRFD,wu2023yunet}, provide very similar results, we empirically selected SCRFD \cite{guo2021SCRFD} to crop facial regions, resizing them to $256 \times 256$ images, and  simultaneously extracting landmark coordinates for the faces in PFED5 and Toronto NeuroFace. 
To mitigate the impact of large head movements and {highlight expression-related variations}, 
all facial landmark coordinates are converted to relative positions 
with respect to the nose tip landmark (see Fig. \ref{fig:split_example} (top)). 

\begin{table*} [ht]
\centering
\caption{{Comparative Spearman’s Rank Correlation results of our method and related SOTA methods on \textbf{PFED5}.
{For LSTPNet \cite{lu2024lstpnet}, since its fine-tuned weights were not available, we trained it on DFEW to obtain the required weights.} All results are in \%. Best result is in bold, second-best is underlined. }}
\label{tab:sota_PD}
\begin{tabular}{m{0.15cm}m{2cm}lm{1.65cm}m{1.45cm}m{0.6cm}ccccc|c}  
\hline
&\textbf{Method} & \textbf{Publication} & \textbf{Backbone} & \textbf{Pre-training} & \begin{tabular}[l]{@{}l@{}}\textbf{Input} \\\textbf{Frames}\end{tabular}& \begin{tabular}[c]{@{}c@{}}\textbf{Sit} \\\textbf{at rest}\end{tabular} & \textbf{Smile} & \textbf{Frown} & \begin{tabular}[c]{@{}c@{}}\textbf{Squeeze} \\\textbf{eyes}\end{tabular} & \begin{tabular}[c]{@{}c@{}}\textbf{Clench} \\\textbf{teeth}\end{tabular} & \begin{tabular}[c]{@{}c@{}}\textbf{Avg. }\\\textbf{Corr. }\end{tabular} \\ 
\hline
\multirow{4}{*}{\raisebox{4ex}{\rotatebox{-90}{DFER}} \raisebox{0.5ex}{$\left\{\rule{0mm}{6.5ex}\right.$}}&Former-DFER\cite{formerDFER} & ACMMM'21 & I3D-ViT & DFEW & 80& 43.77 & 67.32 & 48.89 & 47.88 & 50.91 & 51.75 \\
&GCA-IAL\cite{GCA_IAL} & AAAI'23 & I3D & DFEW &80& 30.83 & \textbf{75.17} & 43.25 & 38.23 & 59.02 & 49.30 \\
&LSTPNet\cite{lu2024lstpnet} & IVT'24 & R18-ViT & DFEW &80& 30.26 & 44.50 & 61.22 & 35.99 & 34.70 & 41.33 \\
&S2D\cite{chen2024static} & TAFFC'24 & \begin{tabular}[c]{@{}l@{}}ViT\\+MobileFaceNet\end{tabular} & \begin{tabular}[c]{@{}l@{}}DFEW\\+300-W\end{tabular} &80+80& 58.23 & 60.04 & 53.62 & 29.90 & 40.16 & 48.39 \\ 
\multirow{12}{*}{\raisebox{5ex}{\rotatebox{-90}{AQA}} \raisebox{2ex}{$\left\{\rule{0mm}{15.5ex}\right.$}}&C3D-LSTM\cite{c3d_lstm_seven} & CVPRW'17 & C3D & UCF-101 &80& 27.62 & 57.46 & 36.36 & 43.59 & 46.69 & 42.34 \\
&USDL\cite{2020USDL} & CVPR'20 & I3D & K-400 &80& 40.08 & 55.90 & 46.36 & 44.45 & 52.17 & 47.79 \\
&CoRe\cite{CoRe} & ICCV'21 & I3D & K-400 &80& 50.23 & 62.99 & 53.57 & 56.57 & 65.05 & 57.68 \\
&AQA-TPT\cite{aqa_tpt} & ECCV'22 & I3D & K-400 &80& 46.27 & 62.32 & 47.87 & 56.18 & \textbf{67.51} & 56.03 \\
&Interpretability\cite{dong2024interpretable} & BMVC'24 & I3D & K-400 &80& 49.91 & 25.99 & 36.15 & 19.96 & 11.88 & 28.78 \\
&CoFInAl\cite{zhou2024cofinal} & IJCAI'24 & I3D & K-400 &80& 36.70 & 45.08 & 38.63 & 24.90 & 40.00 & 37.06 \\
&Interpretability\cite{dong2024interpretable} & BMVC'24 & VST & K-600 &80& 35.19 & 36.69 & 25.02 & 29.23 & 24.00 & 30.03 \\
&CoFInAl\cite{zhou2024cofinal} & IJCAI'24 & VST & K-600 &80& 36.47&	67.17&	37.87&	50.25&	55.35&	49.42  \\ 
&QAFE-Net\cite{duan2023qafe} & WACVW'24 & \begin{tabular}[c]{@{}l@{}}I3D-ViT\\+SlowOnly\end{tabular} & \begin{tabular}[c]{@{}l@{}}DFEW\\+NTU60-XSub\end{tabular} &80+80& \textbf{75.57} & 68.29 & \underline{63.26} & \underline{59.60} & 60.01 & \underline{65.35} \\
 & TraMP-Former (Ours) & - & \begin{tabular}[c]{@{}l@{}}I3D-ViT\\+SkateFormer\end{tabular} & \begin{tabular}[c]{@{}l@{}}DFEW \\+ DFEW\end{tabular} & 80+128 & \underline{73.30} & \underline{72.25} & \textbf{79.01} & \textbf{67.57} & \underline{67.17} & \textbf{71.86} \\
\hline
\end{tabular}
\end{table*}

\noindent {\bf Trajectory Stream --}
SkateFormer \cite{do2024skateformer}, our  trajectory encoder, consists of four blocks with dimensions  $(128,256,256,256)$, respectively. Instead of skeletal joints, trajectory streams are generated on DFEW \cite{DFEW} to pre-train  SkateFormer using the same preprocessing pipeline as above. This also helps to facilitate faster convergence while avoiding overfitting.
Entire DFEW clips were excluded if any frame within them failed to detect a face, often due to low lighting or incomplete faces being in frame.

{For pre-training on DFEW, we set the temporal length $T_{p}=256$. During fine-tuning, however, a shorter temporal length of $T_{p}=128$ is used for PFED5 and Toronto NeuroFace to better align with their specific requirements.
An exception is made for the `Sit at rest' expression in PFED5, where $T_p = 64$ is used due to its shorter clip lengths compared to other expressions. 
Clip reversal 
was randomly applied during training for trajectory augmentation.  We ablate different trajectory lengths in Section \ref{sec:ablation}.} 

\noindent {\bf RGB Stream --}
The Former-DFER \cite{formerDFER} backbone, pre-trained on DFEW, is adopted as the RGB encoder. During training, $T_{f}=80$ frames are randomly sampled from each video, followed by random cropping to $224 \times 224$ and horizontal flipping for data augmentation. For inference, 80 frames are uniformly sampled and resized to $224 \times 224$.

\noindent {\bf Other settings --}
To further enhance training, $5\times$ more iterations per epoch are conducted for both streams. 
The regression head consists of a three-layer MLP with the following layers at $(256,128)$, $(128,64)$ and $(64,1)$. The model is optimised using {SGD}
with a momentum of $0.9$. Training is conducted with a batch size of $4$ for $100$ epochs, starting with an initial learning rate of $0.001$, which is reduced by a factor of $10$ every $40$ epochs.

\subsection{ Comparison with State-of-the-art Methods}
\label{sec:sota_compare}

{We evaluate TraMP-Former on PFED5 and our augmented Toronto NeuroFace datasets, comparing its performance with several SOTA methods\footnote{The comparison with \cite{Szczapa_2022_trajec}, who also used trajectories, is not possible due to the unavailability of their code.} in both DFER and {AQA.}
The DFER methods are adapted by replacing the classification head with a three-layer MLP regression head (i.e., the same as in TraMP-Former).  We use their fine-tuned weights on the DFEW  dataset for initialisation, and {further fine-tune} them on PFED5 and augmented Toronto NeuroFace datasets individually.
For the AQA baselines, given the domain gap between sports actions and facial expressions, }
we directly fine-tune them on PFED5 and augmented Toronto NeuroFace datasets, initialising with their original pre-trained weights.



\noindent {\bf Results on PFED5 Dataset --}
Table \ref{tab:sota_PD} shows the performance of DFER methods 
(\cite{formerDFER}, \cite{GCA_IAL}, \cite{lu2024lstpnet}, \cite{chen2024static}) and
AQA methods 
(\cite{c3d_lstm_seven}, \cite{2020USDL}, \cite{CoRe}, \cite{aqa_tpt}, \cite{dong2024interpretable}, \cite{zhou2024cofinal}, \cite{duan2023qafe}) 
compared to TraMP-Former on the different actions in PFED5 dataset. 
{Our proposed TraMP-Former achieves SOTA average performance using trajectory length $T_{p}=128$ with a correlation of 71.86\%, surpassing QAFE-Net by 6.51\%. It also achieves the highest correlation for `Frown' and `Squeeze eyes', with 79.01\% and 67.57\%, respectively, while ranking second-best for all other expressions.
}

Among the DFER methods, Former-DFER achieves the highest average correlation at 51.75\%, but they are all distinctly below TraMP-Former in performance. The only exception is GCA-IAL \cite{GCA_IAL} which excels in `Smile' at 75.17\% by 2.92\% compared to Tramp-Former  at 72.25\%. Its performance in other actions is otherwise underwhelming. 
{We speculate that the global convolution-attention module in GCA-IAL, which is used to rescale the feature map channels, may enhance features of expressions with broader movement regions (e.g. `Smile' and `Clench teeth') and erroneously suppress features of subtle movements within local facial regions.}

\begin{table*} [t]
\centering
\caption{{Comparative Spearman’s Rank Correlation results on \textbf{augmented Toronto NeuroFace dataset}. 
{For LSTPNet \cite{lu2024lstpnet}, since its fine-tuned weights were not available, we trained it on DFEW to obtain the required weights.} All results are in \%. Best result is in bold, second-best is underlined.}}
\label{tab:sota_toronto}
\begin{tabular}{m{0.15cm}m{2cm}lm{1.65cm}m{1.45cm}m{0.6cm}ccccc|c} 
\hline
&\textbf{Method} & \textbf{Publication} & \textbf{Backbone} & \textbf{Pre-training} &\begin{tabular}[l]{@{}l@{}}\textbf{Input} \\\textbf{Frames}\end{tabular}& \textbf{BLOW} & \textbf{KISS} & \textbf{BROW} & \textbf{OPEN} & \textbf{SPREAD} & \begin{tabular}[c]{@{}c@{}}\textbf{Avg. }\\\textbf{Corr. }\end{tabular} \\ 
\hline
\multirow{4}{*}{\raisebox{4ex}{\rotatebox{-90}{DFER}} \raisebox{0.5ex}{$\left\{\rule{0mm}{5ex}\right.$}}&Former-DFER\cite{formerDFER} & ACMMM'21 & I3D-ViT & DFEW &80& 16.73 & 13.52 & 59.83 & \underline{62.38} & \underline{52.59} & 41.01 \\
&LSTPNet\cite{lu2024lstpnet} & IVT'24 & R18-ViT & DFEW &80& 19.98 & 13.09 & 2.253 & 23.02 & 27.96 & 17.26 \\
&S2D\cite{chen2024static} & TAFFC'24 & \begin{tabular}[c]{@{}l@{}}ViT\\+MobileFaceNet\end{tabular} & \begin{tabular}[c]{@{}l@{}}AffectNet\\+300-W~\end{tabular} &80+80& 4.05 & 3.44 & 42.15 & 26.42 & 19.68 & 19.15 \\
\multirow{8}{*}{\raisebox{4ex}{\rotatebox{-90}{AQA}} \raisebox{0.5ex}{$\left\{\rule{0mm}{10ex}\right.$}}&Interpretability\cite{dong2024interpretable} & BMVC’24 & I3D & K-400 &80& 26.19 & 22.75 & 33.28 & 18.68 & 22.94 & 24.77 \\
&CoFInAl\cite{zhou2024cofinal} & IJCAI'24 & I3D & K-400 &80& 31.24 & 29.13 & 37.72 & 31.59 & 23.81 & 30.70 \\
&Interpretability\cite{dong2024interpretable} & BMVC’24 & VST & K-600 &80& 35.12 & 28.34 & 60.33 & 52.66 & 41.91 & 43.67 \\
&CoFInAl\cite{zhou2024cofinal} & IJCAI'24 & VST & K-600 &80& \underline{35.64} & \textbf{41.05} & 54.54 & 35.23 & 32.15 & 39.72 \\ 
&QAFE-Net\cite{duan2023qafe} & WACVW'24 & \begin{tabular}[c]{@{}l@{}}I3D-ViT\\+SlowOnly\end{tabular} & \begin{tabular}[c]{@{}l@{}}DFEW\\+NTU60-XSub\end{tabular} &80+80& 20.90 & \underline{34.00} & \underline{61.39} & \textbf{64.24} & 50.58 &\underline{46.22} \\
&TraMP-Former (Ours) & - & \begin{tabular}[c]{@{}l@{}}I3D-ViT\\+SkateFormer\end{tabular} & \begin{tabular}[c]{@{}l@{}}DFEW \\+ DFEW\end{tabular}&80+128 & \textbf{44.92} & 30.28 & \textbf{75.23} & 60.22 & \textbf{58.53} & \textbf{53.84} \\
\hline
\end{tabular}
\vspace*{-3mm}
\end{table*}

\noindent {\bf Results on augmented Toronto NeuroFace Dataset --}
We also compare TraMP-Former on the different (non-speech) actions in the augmented Toronto NeuroFace dataset with several SOTA methods, including DFER methods (\cite{formerDFER}, \cite{lu2024lstpnet}, \cite{chen2024static}) and
AQA methods (\cite{dong2024interpretable}, \cite{zhou2024cofinal}, \cite{duan2023qafe}).
{As shown in Table \ref{tab:sota_toronto}, 
TraMP-Former achieves SOTA average performance on this dataset also 
with a Spearman's rank correlation of 53.84\%, which exceeds QAFE-Net by 7.62\%. It also reaches the highest correlation for `BLOW' (44.92\%) and `SPREAD' (58.53\%), outperforming QAFE-Net by margins of 24.02\% and 7.95\%, respectively.}

\begin{table}
\setlength{\tabcolsep}{2.5pt} 
\centering
\scriptsize
\caption{{Comparative classification tasks results on original Toronto NeuroFace dataset. The regression head of TraMP-Former is replaced with a single-layer classification head. Results are in \%. Best result is in bold. 
}}
\label{tab:cls_toronto}
\begin{tabular}{llcccc} 
\toprule
\textbf{ Subscore } & \textbf{ Methods} & \textbf{ Accuracy} & \begin{tabular}[c]{@{}c@{}}\textbf{ Binary }\\\textbf{Accuracy}\end{tabular} & \begin{tabular}[c]{@{}c@{}}\textbf{ Specificity}\\\textbf{(TNR)}\end{tabular} & \begin{tabular}[c]{@{}c@{}}\textbf{ Sensitivity}\\\textbf{(TPR)}\end{tabular} \\ 
\hline
\multirow{2}{*}{ROM} & Ipapo et al. \cite{ipapo2023clinical} & 62.26 & 66.04 & \textbf{94.12} & 15.79 \\
 & TraMP-Former & \begin{tabular}[c]{@{}c@{}}\textbf{66.04}\\(3.78\%↑)\end{tabular} & \begin{tabular}[c]{@{}c@{}}\textbf{81.13}\\(15.09\%↑)\end{tabular} & \begin{tabular}[c]{@{}c@{}}88.41\\(5.71\%↓)\end{tabular} & \begin{tabular}[c]{@{}c@{}}\textbf{44.32}\\(28.53\%↑)\end{tabular} \\ 
\hline
\multirow{2}{*}{Speed} & Ipapo et al. \cite{ipapo2023clinical} & 73.58 & 73.58 & 90.48 & 9.09 \\
 & TraMP-Former & \begin{tabular}[c]{@{}c@{}}\textbf{74.53}\\(0.95\%↑)\end{tabular} & \begin{tabular}[c]{@{}c@{}}\textbf{79.25}\\(5.67\%↑)\end{tabular} & \begin{tabular}[c]{@{}c@{}}\textbf{90.72}\\(0.24\%↑)\end{tabular} & \begin{tabular}[c]{@{}c@{}}\textbf{26.14}\\(17.05\%↑)\end{tabular} \\ 
\hline
\multirow{2}{*}{Symmetry} & Ipapo et al. \cite{ipapo2023clinical} & 58.49 & \textbf{69.81} & \textbf{77.78} & \textbf{61.54} \\
 & TraMP-Former & \begin{tabular}[c]{@{}c@{}}\textbf{59.43}\\(0.94\%↑)\end{tabular} & \begin{tabular}[c]{@{}c@{}}66.98\\(2.83\%↓)\end{tabular} & \begin{tabular}[c]{@{}c@{}}77.15\\(0.63\%↓)\end{tabular} & \begin{tabular}[c]{@{}c@{}}47.22\\(14.32\%↓)\end{tabular} \\
\bottomrule
\end{tabular}
\end{table}

{\noindent {\bf Classification task on Toronto NeuroFace --} To the best of our knowledge, Ipapo et al. \cite{ipapo2023clinical} is the only work which has utilised the score labels from the original Toronto NeuroFace dataset to classify disease severity based on individual subscores. They employed specific geometric features (e.g. Euclidean distance of mouth width) derived from facial landmarks, tailored to individual subscores, and used a Random Forest classifier to predict the scores on a 1 to 4 scale for each feature category.}
Following the same  training (80\%) and test (20\%) split, we evaluated our model against \cite{ipapo2023clinical} 
as shown in Table \ref{tab:cls_toronto}. Our TraMP-Former achieves superior performance across all metrics for ROM and Speed subscores, except for specificity on the ROM subscore. 
We underperform on the Symmetry subscore, 
as Tramp-Former neither incorporates features that can capture symmetry nor does it perform symmetry analysis.


\subsection{Ablations}
\label{sec:ablation}
We perform several ablations on the above two datasets, 
examining the effects of landmark trajectories, TraMP fusion module, and other issues on TraMP-Former's performance. 

{
\noindent {\bf Effect of Landmark Trajectories --} Table \ref{tab:abalation_trajs_input} presents the comparative results of using different input streams: the RGB stream alone, the trajectory stream alone, and the dual-stream setting. 
In Row 2, when engaging only the trajectory-based stream with $(x, y)$ coordinates at each time point, i.e., $C_{p}=2$, the model achieves the highest correlation on `Smile', and outperforms the RGB stream alone (Row 1) for all facial expressions. }
{In Row 3, when using the trajectory stream without positional information and only with their RGB pixel values $(RGB)$ associated with positions over time, i.e., $C_{p}=3$, the model exhibits notable performance degradation. Having spatial coordinates to capture positional relationships between landmarks is clearly important as seen in Row 4 when $C_p=5$, i.e., when the trajectories are represented by $(x,y,RGB)$.
Integrating trajectory features with RGB frame features in our proposed  dual-stream setting significantly improves performance, with overall best results obtained when using $(x,y,RGB)$ in Row 7.}


\begin{table}
\centering
\scriptsize
\caption{ Comparison of different input streams on PFED5.  For the dimension of trajectory input channel $C_{p}$, $C_{p}=2$ represents only using landmark positions $(x, y)$ over time; $C_{p}=3$ indicates only using the RGB pixel values $(RGB)$ associated with $(x, y)$ over time; and $C_{p}=5$ combines $(x, y)$ positions with their corresponding RGB values $(RGB)$ over time, i.e., $(x,y,RGB)$ at each time point. All results are in \%. Best result is in bold, second-best is underlined. $T_p=128$. }
\label{tab:abalation_trajs_input}
\begin{tabular}{>{\centering\arraybackslash}m{0.2cm} >{\centering\arraybackslash}m{0.4cm}c|cccccc} 
\toprule 
\textbf{Row} & \textbf{Stream} & $\boldsymbol{C_{p}}$ & \begin{tabular}[c]{@{}c@{}}\textbf{Sit }\\\textbf{at rest}\end{tabular} & \textbf{Smile} & \textbf{Frown} & \begin{tabular}[c]{@{}c@{}}\textbf{Squeeze }\\\textbf{eyes}\end{tabular} & \begin{tabular}[c]{@{}c@{}}\textbf{Clench }\\\textbf{teeth}\end{tabular} & \begin{tabular}[c]{@{}c@{}}\textbf{Avg.}\\\textbf{Corr.}\end{tabular} \\ 
\hline
1 & RGB & - & 43.77 & 67.32 & 48.89 & 47.88 & 50.91 & 51.75 \\ 
\hline
2 &  & 2 & 51.86 & \textbf{75.87} & 60.12 & 51.75 & 58.51 & 59.62 \\
3 & {Traj.} & 3 & 50.28 & 58.31 & 50.62 & 26.06 & 31.34 & 43.32 \\
4 &  & 5 & 66.54 & 67.75 & 49.53 & 42.27 & 53.94 & 56.01 \\ 
\hline
5 & \multirow{3}{*}{\begin{tabular}[c]{@{}c@{}}RGB\\+Traj.\end{tabular}} & 2 & 68.60 & 70.14 & \underline{75.37} & \underline{63.15} & 65.70 & 68.59 \\
6 & & 3 & \underline{72.71} & \underline{74.27} & 70.37 & 61.03 & \textbf{67.53} & \underline{69.18} \\
7  & & 5 & \textbf{73.30} & 72.25 & \textbf{79.01} & \textbf{67.57} & \underline{67.17} & \textbf{71.86} \\
\bottomrule
\end{tabular}
\end{table}

{\noindent {\bf Fusion strategies --} 
In QAFE-Net \cite{duan2023qafe}, Cross-Fusion is shown to be the best approach to fuse their RGB and landmark heatmaps on PFED5. This approach underperforms when applied to fuse our RGB and trajectory features, which may be due to the inconsistent stream lengths. As shown in Table \ref{tab:abalation_fusion}, our proposed TraMP fusion module addresses this issue by updating only the query stream during the fusion process. Specifically, when the RGB stream is designated as the query (i.e., only the RGB stream is updated), the model efficiently  encodes motion information from trajectory data  and complements it on RGB features.}
\begin{table}
\centering
\scriptsize
\setlength{\tabcolsep}{1.5mm}
\caption{{Comparative results of using TraMP fusion module with other strategies on PFED5. For TraMP (*), * denotes which stream serves as query in TraMP Blocks. Results are in \%,  best in bold, second-best underlined.}}
\label{tab:abalation_fusion}
\begin{tabular}{l|cccccc} 
\toprule
\textbf{Fusion} & \begin{tabular}[c]{@{}c@{}}\textbf{Sit }\\\textbf{at rest}\end{tabular} & \textbf{Smile} & \textbf{Frown} & \begin{tabular}[c]{@{}c@{}}\textbf{Squeeze }\\\textbf{eyes}\end{tabular} & \begin{tabular}[c]{@{}c@{}}\textbf{Clench }\\\textbf{teeth}\end{tabular} & \begin{tabular}[c]{@{}c@{}}\textbf{Avg. }\\\textbf{Corr. }\end{tabular} \\ 
\hline
Summation & 70.24 & 70.36 & \underline{79.11} & \underline{60.57} & 64.45 & 68.95 \\
Concatenation & \underline{71.64} & 71.45 & \textbf{79.56} & 59.47 & 62.70 & \underline{68.96} \\
Cross-Fusion \cite{duan2023qafe} & 70.61 & 69.78 & 71.60 & 55.05 & \textbf{70.05} & 67.42 \\
TraMP (RGB) & \textbf{73.30} & \textbf{72.25} & 79.01 & \textbf{67.57} & \underline{67.17} & \textbf{71.86} \\
TraMP (Trajectory) & 69.09 & \underline{72.23} & 71.74 & 55.72 & 63.86 & 66.53 \\
\bottomrule
\end{tabular}
\end{table}

{\noindent {\bf Number of TraMP Blocks --} 
Table \ref{tab:abalation_num_block} reports the results for the number of TraMP blocks of $1$, $2$, $3$, $4$, and $5$. The model achieves the highest average correlation with 3 blocks, delivering the best performance across all actions except for `Clench teeth', where it ranks second.} 
{Performance tends to drop when the number of blocks exceeds 3, {as too many blocks may cause the cross-attention to gradually ignore the RGB features,} 
which are updated while the trajectory features, used as keys and values, remain unchanged, potentially leading to overfitting.}

\begin{table}
\centering
\scriptsize
\caption{{Effect of using different number of TraMP Blocks in the fusion module  on PFED5. All results are in \%. Best result is in bold, second-best is underlined. }}
\label{tab:abalation_num_block}
\begin{tabular}{c|cccccc} 
\toprule
Blocks & \begin{tabular}[c]{@{}c@{}}\textbf{Sit }\\\textbf{at rest}\end{tabular} & \textbf{Smile} & \textbf{Frown} & \begin{tabular}[c]{@{}c@{}}\textbf{Squeeze }\\\textbf{eyes}\end{tabular} & \begin{tabular}[c]{@{}c@{}}\textbf{Clench }\\\textbf{teeth}\end{tabular} & \begin{tabular}[c]{@{}c@{}}\textbf{Avg.}\\\textbf{Corr.}\end{tabular} \\ 
\hline
1 & \underline{71.34} & \underline{71.35} & 75.37 & 60.36 & 63.69 & 68.42 \\
2 & 70.93 & 68.2 & 77.54 & 58.51 & 63.77 & 67.79 \\
3 & \textbf{73.30} & \textbf{72.25} & \textbf{79.01} & \textbf{67.57} & \underline{67.17} & \textbf{71.86} \\
4 & 68.50 & 66.58 & \underline{77.78} & \underline{65.70} & 65.42 & \underline{68.80} \\
5 & 70.71 & 66.02 & 76.3 & 61.65 & \textbf{68.04} & 68.54 \\
\bottomrule
\end{tabular}
\end{table}

{\noindent {\bf Different types of padding for short clips --} To address shorter clip lengths, we padded the trajectories using looping for both position and RGB values of the trajectory landmarks. In  Table \ref{tab:abalation_padding}, we ablate different padding strategies to justify our approach. Padding by duplication involves repeating the last frame of the clip to match the required length, and padding by interpolation represents cubic splines interpolation for extending the clips. Zero padding indicates that all positions and associated RGB values are set to $0$ for padded frames in the trajectory stream, while in the RGB stream, all RGB pixels are set to $0$. 
It can be seen from the table that interpolation and duplication negatively impact model performance, likely due to altering the original motion speed, which is critical for quality assessment.
In contrast, loop padding for both streams achieves the highest correlation across all expressions except `Smile', where it ranks second among the dual-stream settings.}

\begin{table}
\centering
\scriptsize
\caption{{Effect of different types of padding for RGB and trajectory streams on PFED5.  
{We used $C_{p}=2$ here for the trajectory-only stream  in the first 4 rows due to noise introduced by associated RGB values.} For dual-stream, we used $C_{p}=5$. All results are in \%. Best result is in bold, second-best is underlined.}}
\label{tab:abalation_padding}
\begin{tabular}{p{1.75cm}|cccccc} 
\toprule
\begin{tabular}[c]{@{}l@{}}\textbf{Padding Types}\\\textbf{(RGB/\textbf{Trajectory)}}\end{tabular} & \begin{tabular}[c]{@{}c@{}}\textbf{Sit }\\\textbf{at rest}\end{tabular} & \textbf{Smile} & \textbf{Frown} & \begin{tabular}[c]{@{}c@{}}\textbf{Squeeze }\\\textbf{eyes}\end{tabular} & \begin{tabular}[c]{@{}c@{}}\textbf{Clench }\\\textbf{teeth}\end{tabular} & \begin{tabular}[c]{@{}c@{}}\textbf{Avg.}\\\textbf{Corr.}\end{tabular} \\ 
\hline
--/zero & 46.77 & \textbf{77.33} & 54.53 & 53.12 & 61.65 & 58.68 \\
--/loop & 51.86 & \underline{75.87} & 60.12 & 51.75 & 58.51 & 59.62 \\
--/interpolation & 50.87 & 75.26 & 54.16 & 53.69 & 57.75 & 58.35 \\
--/duplication & 45.88 & 74.35 & 44.58 & 55.00 & 63.42 & 56.65 \\ 
\hline
zero/zero & 69.18 & 68.64 & 73.22 & \underline{62.77} & \underline{65.40} & 67.84 \\
zero/loop & 68.35 & 68.71 & \underline{77.86} & 60.45 & 64.84 & 68.04 \\
loop/zero & \underline{72.65} & 72.77 & 75.44 & 61.99 & 63.04 & \underline{69.18} \\
loop/loop & \textbf{73.30} & 72.25 & \textbf{79.0}1 & \textbf{67.57} & \textbf{67.17} & \textbf{71.86} \\
\bottomrule
\end{tabular}
\end{table}

\noindent {\bf Different trajectory lengths --}
{As SkateFormer allows specific input lengths of $64$, $128$, and $256$, we ablate our trajectory clip length at $T_{p}=64$, $T_{p}=128$, and $T_{p}=256$ on both PFED5 and augmented Toronto NeuroFace datasets (see Table \ref{tab:traj_length}). As shown in the main results Tables \ref{tab:sota_PD} and \ref{tab:sota_toronto}, TraMP-Former  achieves SOTA average performance on both datasets with $T_{p}=128$. When extending the trajectories to the maximum clip length $T_{p}=256$, TraMP-Former's performance still achieves the second-best method on average compared to other methods.  We associate the lower average performance with noise introduced by padding using additional frames since the majority of clips in the dataset are not that long. Additionally,  TraMP-Former with $T_{p}=256$ obtains the best results for `BROW' at 88.15\% and `OPEN' at 68.92\% on augmented Toronto NeuroFace dataset which also ourperforms the methods in Table \ref{tab:sota_toronto}.}

\begin{table}
\centering
\scriptsize
\caption{Comparison of different trajectory lengths on PFED5 (top  3 rows) and augmented Toronto NeuroFace (aTNF, bottom 3 rows). All results are in \%. Best result is in bold, second-best is underlined.}
\label{tab:traj_length}
\begin{tabular}{m{0.10cm}c|cccccc} 
\toprule
&{$\boldsymbol{T_{p}}$} & \begin{tabular}[c]{@{}c@{}}\textbf{Sit} \\\textbf{at rest}\end{tabular} & \textbf{Smile} & \textbf{Frown} & \begin{tabular}[c]{@{}c@{}}\textbf{Squeeze} \\\textbf{eyes}\end{tabular} & \begin{tabular}[c]{@{}c@{}}\textbf{Clench} \\\textbf{teeth}\end{tabular} & \begin{tabular}[c]{@{}c@{}}\textbf{Avg. }\\\textbf{Corr. }\end{tabular} \\ 
\hline
\multirow{3}{*}{\raisebox{4ex}{\rotatebox{-90}{\textbf{PFED5}}} \raisebox{0.5ex}{$\left\{\rule{0mm}{3ex}\right.$}}&64 & \textbf{73.30} & 67.92 & \underline{75.05} & 60.69 & 59.88 & 67.37 \\		
&128 & \textbf{73.30} & \textbf{72.25} & \textbf{79.01} & \textbf{67.57} & \textbf{67.17} & \textbf{71.86} \\
&256 & \underline{70.89} & \underline{69.93} & 71.04 & \underline{63.60} & \underline{63.21} & \underline{67.73} \\ 
\hline
&{$\boldsymbol{T_{p}}$} & \textbf{BLOW} & \textbf{KISS} & \textbf{BROW} & \textbf{OPEN} & \textbf{SPREAD} & \begin{tabular}[c]{@{}c@{}}\textbf{Avg.}\\\textbf{Corr.}\end{tabular} \\ 
\hline
\multirow{3}{*}{\raisebox{4ex}{\rotatebox{-90}{\textbf{aTNF}}} \raisebox{0.5ex}{$\left\{\rule{0mm}{3ex}\right.$}}&64 & \underline{28.52} & 21.16 & 68.76 & \underline{65.06} & 35.60 & 43.82 \\
&128 & \textbf{44.92} & \textbf{30.28} & \underline{75.23} & 60.22 & \textbf{58.53} & \textbf{53.84} \\
&256 & 26.40 & \underline{24.19} & \textbf{88.15} & \textbf{68.92} & \underline{52.60} & \underline{52.05} \\
\bottomrule
\end{tabular}
\end{table}

{\noindent {\bf Extracting trajectories from video data --} Recent tracking methods, such as TAPIR \cite{doersch2023tapir}, CoTracker \cite{karaev2023cotracker}, and BootsTAP \cite{doersch2024bootstap}, have achieved impressive results in tracking keypoints on objects in video data. However, as shown in Table \ref{tab:abalation_source}, SCRFD \cite{guo2021SCRFD}  (a landmark detection method used in this work) outperforms these methods on average (e.g. by a margin of 2.63\% on TAPIR) when applied to spatio-temporal facial landmarks.} 

\begin{table}
\centering
\scriptsize
\caption{{Comparison of using different methods to obtain landmark trajectories from PFED5. All results are in \%. Best result is in bold, second-best is underlined.}}
\label{tab:abalation_source}
\begin{tabular}{l|cccccc} 
\toprule
\textbf{Methods} & \begin{tabular}[c]{@{}c@{}}\textbf{Sit }\\\textbf{at rest}\end{tabular} & \textbf{Smile} & \textbf{Frown} & \begin{tabular}[c]{@{}c@{}}\textbf{Squeeze }\\\textbf{eyes}\end{tabular} & \begin{tabular}[c]{@{}c@{}}\textbf{Clench }\\\textbf{teeth}\end{tabular} & \begin{tabular}[c]{@{}c@{}}\textbf{Avg.}\\\textbf{Corr.}\end{tabular} \\ 
\hline
TAPIR \cite{doersch2023tapir} & \underline{71.07} & 67.88 & 76.41 & 64.14 & \underline{66.6}5 & \underline{69.23} \\
CoTracker \cite{karaev2023cotracker} & 70.15 & \underline{69.10} & 72.34 & 58.79 & 64.22 & 66.92 \\
BootsTAP \cite{doersch2024bootstap} & 69.12 & 67.36 & \underline{77.75} & \underline{66.14} & 65.26 & 69.13 \\ 
\hline
SCRFD \cite{guo2021SCRFD} & \textbf{73.30} & \textbf{72.25} & \textbf{79.01} & \textbf{67.57} & \textbf{67.17} & \textbf{71.86} \\
\bottomrule
\end{tabular}
\end{table}

\section{LIMITATIONS}
\noindent {\bf Large head movements --} Our method mitigates the effects of head translations and forward/backward motion of the
head relative to the camera by cropping facial regions and normalising all landmarks to positions relative to the nose tip {(see Fig. \ref{fig:split_example} (top)). This anchors the facial geometry to a stable reference point, reducing sensitivity to global head motions, and preserving the relative structure of facial features.} However, some noise from large head movements may still negatively affect trajectory accuracy. This limitation stems from relying solely on 2D landmarks for handling head rotations compared to 3D representations.

\noindent {{\bf Limited dataset size --} Although we employ data augmentation and transfer learning to address this limitation of PFED5 to some extent, we also note the difficulties in building large datasets of expressions in patients with neurological disorders. One possible route may be to resort to simulated data \cite{Niinuma_2021_WACV}.}
\section{CONCLUSION}
This work proposed 
using the spatio-temporal dynamics of facial landmark trajectories to complement RGB features for assessing fine-grained facial expression quality. We introduced TraMP-Former, a dual-stream FEQA framework that effectively integrated trajectory and RGB features to capture nuanced facial muscle movement variations from both spatial and temporal perspectives. Extensive experiments on the PFED5 and augmented Toronto NeuroFace datasets demonstrated that TraMP-Former achieved SOTA performance across multiple expressions based on Spearman’s rank correlation, highlighting its robustness and effectiveness in modeling subtle facial dynamics. 
In future work, we aim to incorporate 3D facial landmarks to mitigate {the noise from unexpected head rotations on trajectories} to provide a more accurate representation of facial dynamics within trajectories.


\section{ACKNOWLEDGMENTS}
{
Portions of the research here uses the Toronto NeuroFace Dataset collected by Dr. Yana Yunusova and the Vocal Tract Visualization and Bulbar Function Lab teams at UHN-Toronto Rehabilitation Institute and Sunnybrook Research Institute respectively, financially supported by the Michael J. Fox Foundation, NIH-NIDCD, Natural Sciences and Engineering Research Council, Heart and Stroke Foundation Canadian Partnership for Stroke Recovery and AGE-WELL NCE.}



\section*{ETHICAL IMPACT STATEMENT}
Our research involves video assessments to analyse characteristics of  facial expressions in people with neurological disorders such as Parkinson's Disease. 

\noindent \textbf{Ethical Review Boards --}
We signed the license agreements to access PFED5 and Toronto NeuroFace datasets separately, which are publicly available to those who agree to the licences. The datasets have been cleared for public release by their respective Ethics boards. 


\noindent \textbf{Potential Harms to Human Subjects --}
To the best of our knowledge, no harm came to the subjects recorded in the two datasets. 



\noindent \textbf{Potential Negative Societal Impacts --}
The widespread adoption of AI-based technologies raises privacy concerns, as individuals may not want their health conditions to be detected or inadvertently disclosed through casual video. While our primary goal is to assist in the severity assessment of neurological disorders (e.g., Parkinson's Disease), this technology is intended to complement, not replace, professional diagnoses by healthcare providers.

\noindent \textbf{Risk-Mitigation Strategies --} In addition to compliance with ethical review requirements, we access data in accordance with the corresponding license agreements, which explicitly prohibit redistribution and commercial use. All data accessed is securely hosted on a dedicated platform to ensure confidentiality and integrity.

There are no conflicts of interest among the contributors to this study.


{\small
\bibliographystyle{ieee}
\bibliography{main_bib}
}

\end{document}